\documentclass{sig-alternatereducedauth}

\usepackage{url,amssymb,amsfonts,amsmath}
\usepackage{algorithm}
\usepackage[noend]{algpseudocode}
\usepackage{graphicx}
\usepackage{pdfpages}
\usepackage{caption}
\usepackage{multirow}
\usepackage{xcolor,colortbl}
\usepackage{cite}
\usepackage[update,prepend]{epstopdf}

\DeclareGraphicsExtensions{.pdf,.png,.jpg}

\begin{document}

\setlength{\pdfpagewidth}{8.5in}
\setlength{\pdfpageheight}{11in}

\title{Evaluation of Explore-Exploit Policies in Multi-result Ranking Systems}

\numberofauthors{3} %  in this sample file, there are a *total*

\author{
\alignauthor
%Blind Submission
Dragomir Yankov\\
       \email{dragoy@microsoft.com}
% 2nd. author
\alignauthor
Pavel Berkhin\\
       \email{pavelbe@microsoft.com}
% 3rd. author
\alignauthor Lihong Li\\
       \email{lihongli@microsoft.com}
}

% There's nothing stopping you putting the seventh, eighth, etc.
% author on the opening page (as the 'third row') but we ask,
% for aesthetic reasons that you place these 'additional authors'
% in the \additional authors block, viz.
%\additionalauthors{Additional authors: John Smith (The Th{\o}rv{\"a}ld Group,
%email: {\texttt{jsmith@affiliation.org}}) and Julius P.~Kumquat
%(The Kumquat Consortium, email: {\texttt{jpkumquat@consortium.net}}).}
%\date{\today}
% Just remember to make sure that the TOTAL number of authors
% is the number that will appear on the first page PLUS the
% number that will appear in the \additionalauthors section.

\maketitle
\begin{abstract}

We analyze the problem of using Explore-Exploit techniques to improve precision in multi-result ranking systems such as web search, query autocompletion and news recommendation. Adopting an exploration policy directly online, without understanding its impact on the production system,  may have unwanted consequences - the system may sustain large losses, create user dissatisfaction, or collect exploration data which does not help improve ranking quality. An offline framework is thus necessary to let us decide what policy and how we should apply in a production environment to ensure positive outcome. Here, we describe such an offline framework.

Using the framework, we study  a popular exploration policy --- Thompson sampling. We show that there are different ways of implementing it in multi-result ranking systems, each having different semantic interpretation and leading to different results in terms of sustained click-through-rate (CTR) loss and expected model improvement. In particular, we demonstrate that Thompson sampling can act as an online learner optimizing CTR, which in some cases can lead to an interesting outcome: lift in CTR during exploration. The observation is important for production systems as it suggests that one can get both valuable exploration data to improve ranking performance on the long run, and at the same time increase CTR while exploration lasts.

\end{abstract}

%A category including the fourth, optional field follows...
%\category{D.2.8}{Software Engineering}{Metrics}[]

 % NOT required for Proceedings

\section{Introduction}

\begin{figure*}[!t]%
\centering
	\includegraphics[width=6.6in]{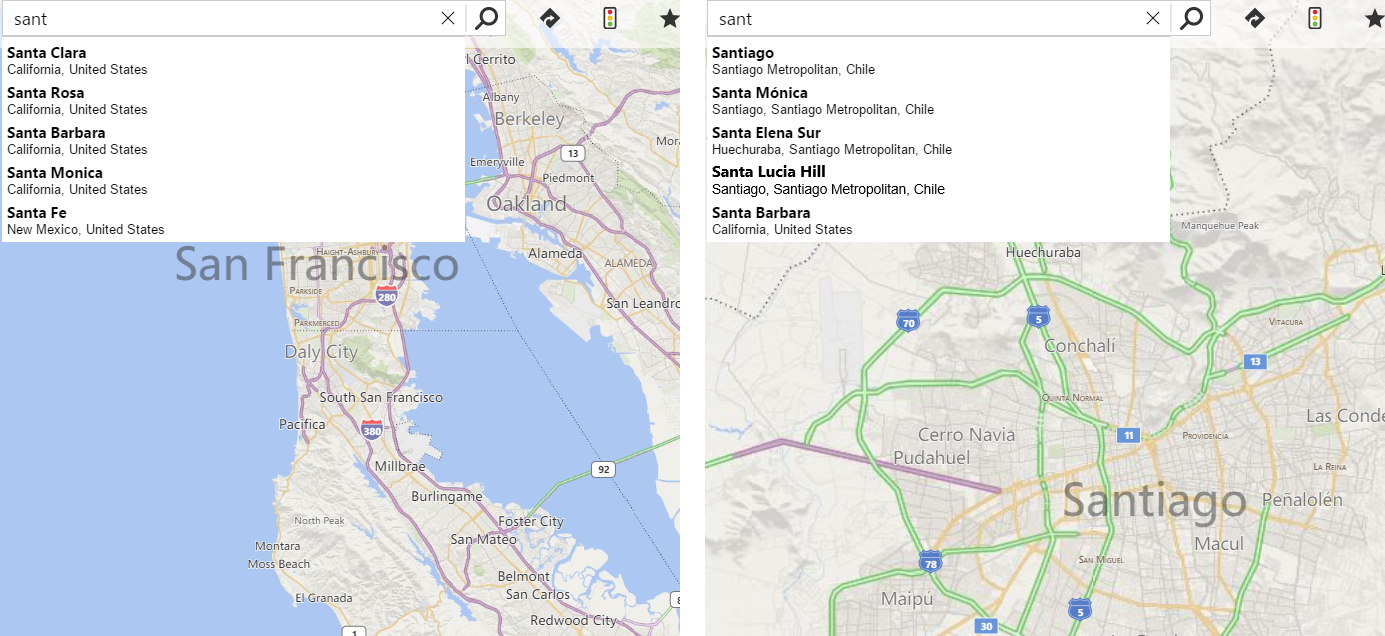}
\caption{Maps query autocomplete system. For the same (query) prefix ``santa'' the system ranks on top geo-results which it deems relevant to the user context. \emph{Left}: San Francisco, \emph{Right}:  Santiago.}
\label{fig:autosuggest} % caption for the whole figure
\end{figure*}

We study ``multi-result'' ranking systems, i.e., systems which rank a number of candidate results and present the top $N$ to the user. Examples of such systems are web search, query autocompletion (see Figure~\ref{fig:autosuggest}), news recommendation, etc. This is in contrast to ``single-result'' ranking systems which also internally utilize ranking mechanisms but in the end display only one result to the user.

One challenge with ranking systems in general is their counterfactual nature~\cite{Bottou_Counterfactual2013}: We cannot directly answer questions of the sort ``Given a query, what would have happened if we had shown a different set of results?" as this is counter the fact. The fact is that we showed whatever results the current production model considered best. Learning new models is thus biased and limited by the deployed ranking model. One sound and popular approach to breaking the dependence on an already deployed model is to integrate an Explore-Exploit (EE) component into the production system~\cite{Li14Counterfactual}. Exploration allows for occasionally randomizing the results presented to the user by overriding some of the top choices of the deployed model and replacing them with potentially suboptimal results. This leads to collecting in our data certain random results generated with small probabilities. When training subsequent ranking models, these results are often assigned higher weights, inversely proportional to the probabilities with which they were explored \cite{Bottou_Counterfactual2013,Strehl_LoggedData2011}.  As theoretically justified and empirically demonstrated, exploration usually allows better models to be learned.  However, adopting exploration in a production system prompts a set of essential questions: which EE policy is most suitable for the system; what would be the actual cost of running EE; and most importantly, how to best use the exploration data to train improved models and what improvements are to be expected?

Here, we present an offline framework which allows ``replaying'' query logs to answer counterfactual questions. The framework can be used to answer the above exploration questions, allowing one to compare different EE policies prior to their integration in the online system. This is very important as running an inadequate policy online can quickly lead to significant money loss, cause broad user dissatisfaction, or collect exploration data which is altogether useless in improving the ranking model.   

As a concrete example, we use the offline framework to evaluate Thompson sampling, a popular EE method which is simple to implement and is very effective at trading off exploration and exploitation~\cite{Chapelle_Thompson11,Scott10Modern,Thompson33Likelihood}. We point out that in fact there are multiple ways of implementing Thompson sampling, each having different semantic interpretation. Some of the implementations correct for bias (calibration problems) in the ranking model scores while others correct for position bias in the results. Naturally, employing different strategies leads to different costs, i.e. the price to be paid for exploring suboptimal results, and to different model improvements. We also introduce two schemes for weighting of examples, collected through exploration, during training new ranking models. 

Because EE can promote suboptimal results it is commonly presumed that production systems adopting it always sustain a drop in the click-through-rate (CTR) during the period of exploration. By analyzing Thompson sampling policies through our offline evaluation framework, we observe an interesting phenomenon: using the right implementation can, in fact, produce a lift in CTR of the production system. In other words, the gain is twofold - the system collects valuable training data and has an improved CTR while exploration continues.

To summarize, this paper makes the following contributions:
%some of the main points introduced in the paper:
\begin{itemize}
\item We describe a novel framework for offline evaluation and comparison of explore-exploit policies in multi-result ranking systems.
\item We introduce several new Thompson sampling implementations that are tailored to multi-result ranking systems, some more suitable in the case of ranking score bias and others in the case of position bias.
\item We introduce two simple weighting schemes for training examples collected through Thompson sampling.
\item Using the framework, we show that adopting the right policy can achieve exploration and increase the CTR of the production system during exploration. 
\end{itemize}

The rest of the paper is organized as follows.  In Section~\ref{sec:framework}, we introduce the offline policy evaluation framework using a maps query autocompletion system as an example. Section~\ref{sec:thompson} discusses different ways of implementing Thompson sampling and weighting examples collected with them. Section~\ref{sec:evaluation} compares the introduced Thompson sampling policies. We finish by discussing related works in Section~\ref{sec:relatedwork}.

%%%%%%%%%%%%%%%%%%%%%%%%%%%%%%%%%%%%%%%%%%%%%%%%%%%%%%%%%%%%%%%%%
% Section: System
%%%%%%%%%%%%%%%%%%%%%%%%%%%%%%%%%%%%%%%%%%%%%%%%%%%%%%%%%%%%%%%%%
\section{Explore-exploit framework} \label{sec:framework}
\subsection{A multi-result ranking system}
As a working example we are going to look into the maps query autocompletion service of a popular map search engine (Figure~1). When users start typing a ``query'' in the system they are presented with up to $N=5$ relevant \emph{geo entities} as suggestions. If users click on one of the results, we assume that we have met their intent; if they do not, the natural question to ask is ``Could we have shown a different set of results which would get a click?''. As pointed out in the introduction, the question is counterfactual and cannot be answered easily as it requires showing a different set of suggestions on exactly the same context. This section describes a framework that allows answering such counterfactual questions. We first go over some prerequisites.

In building ranking systems as above, the usual process goes roughly through the following three stages: 1) The query is matched against an index; 2) For all matched entities a first layer of ranking,  $L_1$ ranker, is applied. The goal of the $L_1$ ranker is to ensure very high recall and prune the matched candidates to a more manageable set of say a few hundred results; 3) A second layer, $L_2$ ranker, is then applied which re-orders the $L_1$ results in a way to ensure high precision. There could be more layers with some specialized functionality but overall these three stages cover the three important aspects: matching, recall and precision. 

Building the $L_1$ ranker is beyond the scope of this work. Here, we focus on methodologies for improving $L_2$ ranking, namely, precision of the system. We assume that there is a machine learned model powering the $L_2$ ranker, i.e. the system adopts a learning-to-rank approach~\cite{ChapelleLearningToRank}. 

Let us now focus on the structure of the logs generated by the system as well as how an EE policy can be applied in it. Table~\ref{tab:logs} shows what information can be logged by a multi-result ranking system for a query.

\begin{table} [h]
\begin{tabular}{|r|c|c|r|}
\hline
Position ($i$)&Label ($y$)&Result($r$)&Rank score ($s$) \\
\hline \hline
$i=1$&0&Suggestion 1 &$s_1=0.95$ \\
$i=2$&0&Suggestion 2 &$s_2=0.90$ \\
$i=3$&1&Suggestion 3 &$s_3=0.60$ \\
$i=4$&0&Suggestion 4 &$s_4=0.45$ \\
$i=5$&0&Suggestion 5 &$s_5=0.40$ \\
\hline
\end{tabular}
\caption{Original system: example logs for one query.}\label{tab:logs}
\end{table}

Suppose that $L_1$ ranking extracted $M\geq N$ relevant results which then $L_2$ re-ranked and produced the top $N=5$ suggestions from Table~\ref{tab:logs}. We assume that, for at least a fraction of the queries, the suggestions from the table and the user actions are logged by the production system. 

The first column in the table shows the ranking position of the suggested result. The \emph{Label} column reflects the observed clicks --- $1$ if the result was clicked and $0$ otherwise. The \emph{Result} column contains some context about the result that was suggested, part of which is only displayed to the user and the rest is used to extract features to train the ranking model. The last column shows the score which the $L_2$ ranking model has assigned to the results. 

\subsection{Explore-Exploit in the online environment} \label{sec:EEonline}

We assume a relatively conservative exploration process taking place in the online environment. Namely, it is allowed to replace only suggestions appearing at position $i=N$. This is to ensure that we do not generate large user dissatisfaction by placing potentially bad results as top suggestions. For exploration in addition to the candidate at position $i=N$ we choose among the candidates which $L_1$ returns and $L_2$ ranks at positions $i=N+1,\ldots,i=N+t\leq M$, for some relatively small $t$. By doing so, we do not explore results which are very low down the ranking list of $L_2$ as they are probably not very relevant. Requiring that a candidate for exploration meets some minimum threshold for its ranking score is also a good idea. Naturally, if for a query there are less than $N=5$ candidates, then no exploration takes place for it.

To enable EE online we also need to define a \emph{policy} --- a mechanism which selects with a certain probability a suggestion different from the one that the deployed ranking model would recommend. Different policies can be implemented. In Section~\ref{sec:thompson} we study several such policies and how they can be simulated in an offline environment. We also discuss how the data collected from them can be weighted suitably for training better ranking models; details can be found in Section~\ref{sec:weighting}.

\subsection{Offline simulation environment} \label{sec:simulation}
Running the above EE process directly in the production environment can lead to costly consequences: it may start displaying inadequate results which can cause the system to sustain significant loss in CTR in a very short time. It is further unclear whether it will help us collect training examples that will lead to improving the quality of the ranking model. We therefore want to simulate the above online process a priori in an offline system that closely approximates it. Here, we present such an offline system. 

The main idea of the offline system is to mimic a scaled-down version of the production system.  Specifically, we assume offline that our Autocomplete system displays $k<N$ results to the user instead of $N=5$. Again, to replicate the online EE process from Section~\ref{sec:EEonline}, different policies evaluated in the offline system will be allowed to show on its last position (i.e., on position $k$) any of the results from positions $i=k,...,N$. 

To understand the offline process better, let us look into two concrete instantiations of the simulation environment which use the logged results from Table~\ref{tab:logs}.

In the first instantiation, we set $k=2$. It means that the offline system displays to the user two suggestions, as seen in Table~\ref{tab:logsk1}. Position $i=2$ is going to be used for exploration, and the result to be displayed will be selected among the candidates at position $i=2,\ldots,i=5$. 
\begin{table} [h]
\begin{tabular}{c|c|c|c|c|l}
\cline{2-5}
& Pos &Label&Result&Score($s$)  &\\ 
\cline{1-5}
\multicolumn{1}{ |r|  }{\multirow{1}{*}{Dis-} } & $i=1$&0&Sugg. 1 &$s_1=0.95$  &   \multicolumn{1}{ c  }{}\\ 
\cline{2-6}
\multicolumn{1}{ |c|  }{\multirow{1}{*}{played} } &\cellcolor{blue!25}$i=2$&\cellcolor{blue!25}0&\cellcolor{blue!25}Sugg. 2 &\cellcolor{blue!25}$s_2=0.90$ &\multicolumn{1}{ c|  }{\multirow{2}{*}{EE Can-}}\\ 
\cline{1-5}
\multicolumn{1}{ c|  }{}                        &$i=3$&1&Sugg. 3 &$s_3=0.60$ & \multicolumn{1}{ l|  }{\multirow{2}{*}{didates}}\\ 
\cline{2-5}
\multicolumn{1}{ c|  }{}                        &$i=4$&0&Sugg. 4 &$s_4=0.45$ &  \multicolumn{1}{ c|  }{}\\ 
\cline{2-5}
\multicolumn{1}{ c|  }{}                        &$i=5$&0&Sugg. 5 &$s_5=0.40$ & \multicolumn{1}{ c|  }{}\\ 
\cline{2-6}
\end{tabular}
\caption{Offline system with $k=2$. Logs for one query derived from the logs from Table~\ref{tab:logs}.  Position $i=2$ (in blue) is used for exploration. }\label{tab:logsk1}
\end{table}

In the second instantiation, we set $k=3$; that is, the offline system is assumed to display three suggestions. Position $i=3$ is used for exploration and the candidates for it are the results from the original logs at positions $i=3,4,5$. This setting is depicted in Table~\ref{tab:logsk3}.
\begin{table} [h]
\begin{tabular}{c|c|c|c|c|l}
\cline{2-5}
& Pos &Label&Result&Score($s$) & \\ 
\cline{1-5}
\multicolumn{1}{ |r|  }{\multirow{2}{*}{Dis-} } & $i=1$&0&Sugg. 1 &$s_1=0.95$     & \\ 
\cline{2-5}
\multicolumn{1}{ |c|  }{\multirow{2}{*}{played} }                        &$i=2$&0&Sugg. 2 &$s_2=0.90$ & \\ 
\cline{2-6}
\multicolumn{1}{ |c|  }{ }                        &\cellcolor{blue!25}$i=3$&\cellcolor{blue!25}1&\cellcolor{blue!25}Sugg. 3 &\cellcolor{blue!25}$s_3=0.60$  &   \multicolumn{1}{ c|  }{\multirow{2}{*}{EE Can-}}\\ 
\cline{1-5}
\multicolumn{1}{ c|  }{\multirow{2}{*}{} } &$i=4$&0&Sugg. 4 &$s_4=0.45$  &   \multicolumn{1}{ l|  }{\multirow{2}{*}{didates}} \\ 
\cline{2-5}
\multicolumn{1}{ c|  }{}                        &$i=5$&0&Sugg. 5 &$s_5=0.40$  &   \multicolumn{1}{ l|  }{} \\ 
\cline{2-6}
\end{tabular}
\caption{Offline system with $k=3$ suggestions. Logs for one query derived from the logs from Table~\ref{tab:logs}. Position $i=3$ (in blue) is used for exploration.}\label{tab:logsk3}
\end{table}

Suppose we use $k=2$. Using only the production system we would display in our simulated environment ``Suggestion 1'' and ``Suggestion 2'' and we would not observe a click as the label in the logs for both position $i=1$ and $i=2$ is zero. Now suppose we use the described framework to compare two EE policies, $\pi_1$ and $\pi_2$, each selecting a different result to display at position $i=2$. For example, $\pi_1$ can select to preserve the result at position $i=2$ (``Suggestion 2'') while $\pi_2$ can select to display instead the result at position $i=3$ (``Suggestion 3''). Now we can ask the counterfactual, with respect to the simulated system,  question ``What would have happened had we applied either of the two policies?''. The answer is, with $\pi_2$ we would have observed a click, which we know from the original system logs (Table~\ref{tab:logs}) and with $\pi_1$ we would not have. If this is the only exploration which we perform the information obtained with $\pi_2$ would be more valuable and would probably lead to training a better new ranking model. Note also that applying $\pi_2$ would actually lead to a higher CTR than simply using the production system. This gives an intuitive idea of why CTR can increase during exploration as demonstrated in the evaluation in Section~\ref{sec:evalpolicies}.

It should be noted that our simulation environment effectively assumes the same label for an item when it is moved to position $k$ from another, lower position $k'>k$.  Due to position bias, CTR of an item tends to be smaller if the item is displayed in a lower position.  Therefore, our simulation environment has a one-sided bias, favoring the production baseline that collects the data.  While the bias makes the offline simulation results less accurate, its one-sided nature implies the results which we show in Section~\ref{sec:evalpolicies} are conservative: if a new policy is shown to have a higher offline CTR in the simulation environment than the production baseline, its online CTR can only be higher in expectation.

%%%%%%%%%%%%%%%%%%%%%%%%%%%%%%%%%%%%%%%%%%%%%%%%%%%%%%%%%%%%%%%%%
% Section: Thompson sampling
%%%%%%%%%%%%%%%%%%%%%%%%%%%%%%%%%%%%%%%%%%%%%%%%%%%%%%%%%%%%%%%%%
\section{Thompson sampling for multi-result ranking}\label{sec:thompson}

We now demonstrate the offline framework by comparing several implementations of Thompson sampling, a popular policy due to its simplicity and effectiveness~\cite{Chapelle_Thompson11,Scott10Modern,Thompson33Likelihood}. In the process we identify two interesting observations. First, there are multiple ways to implement Thompson sampling for multi-result ranking problems. They have different interpretations and lead to different results. Second, if the ``right'' implementation for the problem at hand is selected, then Thompson sampling can refine the ranking model CTR estimates to yield better ranking results.  %for certain examples. 
The method then essentially works as an online learner, improving the CTR of the underlying model by identifying segments where the model is unreliable and overriding its choice with a better one. This in turn can lead to an important result --- adopting EE can entail a twofold benefit: it can collect valuable data for improving ranking precision, and at the same time lift the CTR of the production system during the period of exploration.

Algorithm~\ref{alg:thompson} outlines our \emph{generic}  implementation of Thompson sampling. The algorithm closely follows the implementation suggested by~\cite{Chapelle_Thompson11} with two subtle modifications: 1) defining exploration \emph{buckets}\footnote{Aka ``arms'' in the multi-arm bandits literature.} in line 1; and 2) sampling from the subset of buckets relevant only to the current iteration in line 4. We elaborate on these points below, and will show they are essential and lead to very different results with different instantiations. In the evaluation section we also discuss how we set the exploration constant $\epsilon$ (lines 9, 11) to update the parameters of the beta distributions.

\begin{algorithm}[!t]
\caption{Thompson Sampling for Multi-result Ranking}\label{alg:thompson}
\begin{algorithmic}[1]
\State Define buckets: $\mathbb{P}=\{P_1, P_2,\ldots, P_n\}$ 
\State Initialize Beta distributions: $\{B(\alpha_1,\beta_1),\ldots,B(\alpha_n,\beta_n)\}$  \Comment{One per bucket}
\While{($ExplorationIsEnabled$)} 
	\State $\mathbb{\hat{P}} \gets \{P_{c_1},\ldots,P_{c_l}\} \subseteq \mathbb{P}$ \Comment{Select the buckets involved in the current iteration}
	\State Draw $\theta_i \sim B(\alpha_i, \beta_i), \forall i \in \{c_1, \ldots, c_l\}$ 
	\State $m\gets argmax_i \theta_i$
	\State Display $r_m$ for exploration \Comment{$r_m$ is the result associated with the selected bucket $P_m$}
	\If{($r_m$ clicked)}
	\State $\alpha_m \gets \alpha_m +\epsilon$  \Comment{Increment with constant}
	\Else
	\State $\beta_m \gets \beta_m +\epsilon$     \Comment{Increment with constant}
	\EndIf
\EndWhile
\end{algorithmic}
\end{algorithm}

\subsection{Thompson sampling policies} \label{sec:policies}
Here we describe three policies based on Thompson sampling. Each is characterized by: 1) how it  defines the buckets (line 1 in the algorithm); 2) What probability estimate is the bucket definition semantically representing.
\paragraph{Sampling over positions policy}
This is probably the most straight-forward, but not very effective implementation of Thompson sampling. It defines buckets over the ranking positions used for drawing exploration candidates. More specifically we have: \\
\textbf{Bucket definition:} There are $n = N-k+1$ buckets each corresponding to one of the candidate positions $i=k, ..., i=N$. All of them can be selected in each iteration so $\mathbb{\hat{P}} = \mathbb{P}$. For instance, if we have instantiation as per Table~\ref{tab:logsk3} we would have three buckets, $n=3$, for positions $i=3$, $i=4$, and $i=5$ --- the positions of the candidates for exploration.\\
\textbf{Probability estimate:} $P(click|i,k)$. In this implementation Thompson sampling estimates the probability of click given that a result from position $i$ is shown on position $k$. This implementation allows for correction in the estimate of CTR per position. The approach allows also for correcting position bias. Indeed, results which are clicked simply because of their position may impact the ranking model and during exploration we may through them away eliminating their effect on the system. This makes the approach especially valuable in systems with pronounced position bias. 

\paragraph{Sampling over scores policy}
In this implementation we define the buckets over the scores of the ranking model. Each bucket covers a particular score range. For simplicity let us assume that the score interval $[0,1]$ is divided into one hundred equal subintervals one per each percentage point: $[0, 0.01), [0.01,0.02),\ldots,[0.99,1]$. For the suggested division we have:\\
\textbf{Bucket definition:} There are $n=100$ buckets one per score interval $P_1=[0,0.01),\ldots,P_{100}=[0.99,1]$. In each iteration only a small subset of these are active. In the example from  Table~\ref{tab:logsk3} only the following three buckets are active $P_{c_1}=P_{61}=[0.60,0.61)$, $P_{c_2}=P_{46}=[0.45,0.46)$, and $P_{c_3}=P_{41}=[0.40,0.41)$. Suppose after drawing from their respective Beta distributions we observe that $m=61$, i.e. we should show the first of the three candidates which turns out to result in a click. In this case, we update the positive outcome parameter for the corresponding Beta to $\alpha_{61} = \alpha_{61}+\epsilon$.\\
\textbf{Probability estimate:} $P(click|s,k)$. In this implementation Thompson sampling estimates the probability of click given a ranking score $s$ for a result when shown at position $k$. In general, if we run a calibration procedure then the scores are likely to be close to the true posterior click probabilities for the results~\cite{NiculescuMizilCalibration2005}, but this is only true if we look at them agnostic of position. With respect to position $i=k$ they may not be calibrated. We can think of Thompson sampling as a procedure for calibrating the scores from the explored buckets to closely match the CTR estimate with respect to position $i=k$.

\paragraph{Sampling over scores and positions policy}
This is a combination of the above two implementations. Again we assume that the score interval is divided into one hundred equal parts $[0, 0.01), [0.01,0.02),\ldots,[0.99,1]$. This, however, is done for each candidate position $i=k,\ldots,i=N$.  That is, we have:\\
\textbf{Bucket definition:} $(N-k+1)*100$ buckets. For more compact notation let us assume that bucket $P^i_q$ covers entities with score in the interval $[s, s+0.01)$ when they appear on position $i$ (here $q=\lfloor100s\rfloor+1$). In the example from Table~\ref{tab:logsk3} we have $n=300$ buckets and for the specific iteration the three buckets to perform exploration from are $P^3_{61}$, $P^4_{46}$, and $P^5_{41}$.\\
\textbf{Probability estimate:} $P(click|s,i,k)$. In this implementation Thompson sampling estimates the probability of click given a ranking score $s$ and original position $i$ for a result when it is shown at position $k$ instead. This differs from the previous case as in its estimate it tries to take into account the position bias, if any, associated with clicks. 

These are definitely not all buckets that could be defined. Depending on the concrete system there may be others that are more suitable and lead to even better results. The point we are trying to convey is that how one defines buckets can vary and it plays a crucial role for the success of the exploration process.

\subsection{Example weighting} \label{sec:weighting}
Once we have an EE procedure in place a natural question to ask is how to best use the exploration data to train improved models. Here we introduce two schemes for assigning training weights to examples collected through exploration.
\begin{figure}[!t]
\centering
	\includegraphics[width=3.35in,bb=100 240 485 550, clip]{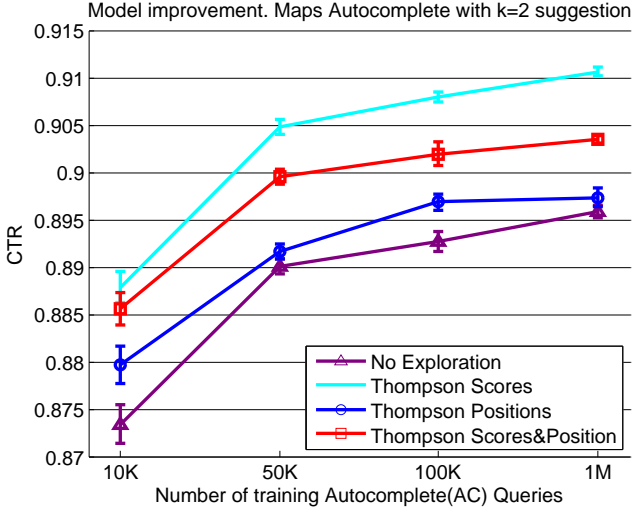}
\caption{Model improvement for an Autocomplete system displaying $k=2$ results to the user.}
\label{fig:modelimprovementk2} 
\end{figure}

\paragraph{Propensity based weights}
In training new rankers it is a common practice to re-weight examples selected through exploration inversely proportional to the probability of displaying them to the user. The probability of selecting an example for exploration is called \emph{propensity score}~\cite{Li14Counterfactual}. More specifically, if we denote the propensity score for example $x_j$ with $p(x_j)$, then its training weight is set to $w_j = \frac{1}{p(x_j)}$. 

Computing propensity scores for Thompson sampling in the general case of more than two Beta distributions involved in each iteration ($l>2$, line 5 in Algorithm~\ref{alg:thompson}) does not have an analytical solution~\cite{Cook_ProbInequalities2008}. We thus derive the following empirical estimate: if we draw for exploration $x_j$ from bucket $P_i$ (line 6 of the algorithm) then we set $p(x_j)$ to the ratio between examples that have been explored thus far from $P_i$ over the sum of all examples explored from buckets in $\mathbb{\hat{P}}$.

\paragraph{Multinomial weights}
We also analyze a different, very simple weighting scheme based on the scores of the baseline ranking model. Let $x_i$ is the result displayed to the user from bucket $P_i$ and let its ranking score be $s_i$. If $x_j$ is the selected example for exploration then we first compute the ``multinomial probability'' $p(x_j) = \frac{s_j}{\sum_{ i \in \{c_1,\ldots,c_l\}}{s_{i}}}$. The weight is then computed again as the inverse proportional $w_j = \frac{1}{p(x_j)}$. If in Table~\ref{tab:logsk3} we have selected for exploration the example at position $i=3$ then its probability is $\frac{0.6}{0.6 + 0.45 + 0.40}=\frac{0.6}{1.45}$ and the weight is $\frac{1.45}{0.6} = 2.41$. We call this weighting scheme \emph{multinomial weighting}.

In both weighting schemes we cap the weight assigned to an example to avoid stressing excessively a single piece of evidence as suggested in previous work~\cite{Bottou_Counterfactual2013,Li14Counterfactual,Strehl_LoggedData2011}.

%%%%%%%%%%%%%%%%%%%%%%%%%%%%%%%%%%%%%%%%%%%%%%%%%%%%%%%%%%%%%%%%%
% Section: Evaluation
%%%%%%%%%%%%%%%%%%%%%%%%%%%%%%%%%%%%%%%%%%%%%%%%%%%%%%%%%%%%%%%%%
\section{Policy Evaluation and Comparison}\label{sec:evaluation}

\begin{figure}[!t]
\centering
	\includegraphics[width=3.35in,bb=100 240 485 550, clip]{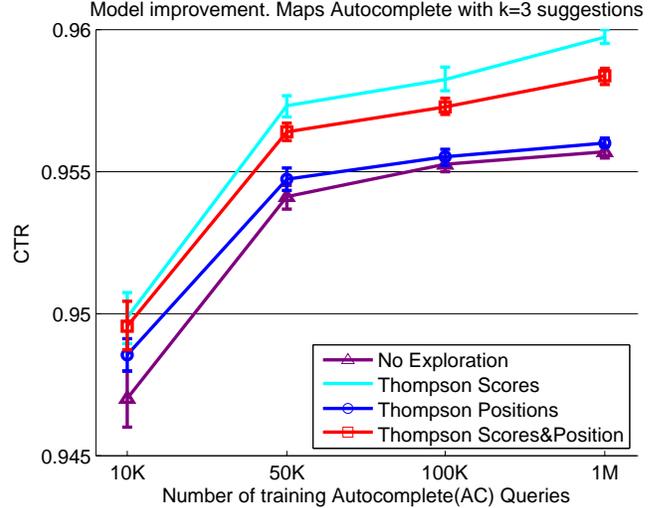}
\caption{Model improvement for an Autocomplete system displaying $k=3$ results to the user.}
\label{fig:modelimprovementk3} 
\end{figure}

The proposed offline framework allows us to evaluate different EE policies, such as the Thompson sampling policies described in the previous section. We mainly focus on two aspects: 1) Expected ranking model improvement; 2) Change in CTR during exploration.  We start with our experimental setup.
%Before we delve into them we explain our setup.

\subsection{Evaluation setup}
\textbf{Data and Baseline Model}: We use maps autocomplete (AC) logs spanning over two months collected by the production model. The first month is used to simulate online exploration, as described in Section~\ref{sec:simulation}, to compute the expected CTR loss and subsequently to train new ranking models. The second month is used to test and report performance results for the models trained on data from the first month. 

Our baseline, \emph{No Exploration} method (Figures~\ref{fig:modelimprovementk2} and~\ref{fig:modelimprovementk3}), closely replicates the $L_2$ ranking model used in the production system --- a variant of boosted regression trees~\cite{Friedman00greedyfunction}. The model adopts a point-wise learning-to-rank approach, that is it is trained on individual suggestions and assigns a clickability score to each result in  scoring time optimizing squared loss. Point-wise models have been shown to be very competitive on a variety of ranking problems~\cite{ChapelleLearningToRank}. It is important to note, though, that the ideas presented here are agnostic to the exact $L_2$ ranker deployed in production. For instance, we obtained directionally similar %to the presented 
results with more elaborate pairwise rankers optimizing NDCG~\cite{WuBoostingIR2010}. 
 
\textbf{Evaluation process}: 
We measure the performance for datasets of sizes 10K, 50K, 100K and 1M queries which we sample uniformly at random from the training month. Keep in mind, that each query leads to up to five suggestions, i.e. for the largest set of 1M queries we train on over 2M individual examples. Each example is represented as a high dimensional vector of features capturing information about the result and the user context --- static rank of the result, similarity to the query, distance to the user, etc.

For each of the query sets we run ten times the following steps, summarizing in the figures the mean and the variance across the ten runs.

\begin{figure}[!t]
\centering
	\includegraphics[width=3.2in,bb=133 250 480 530, clip]{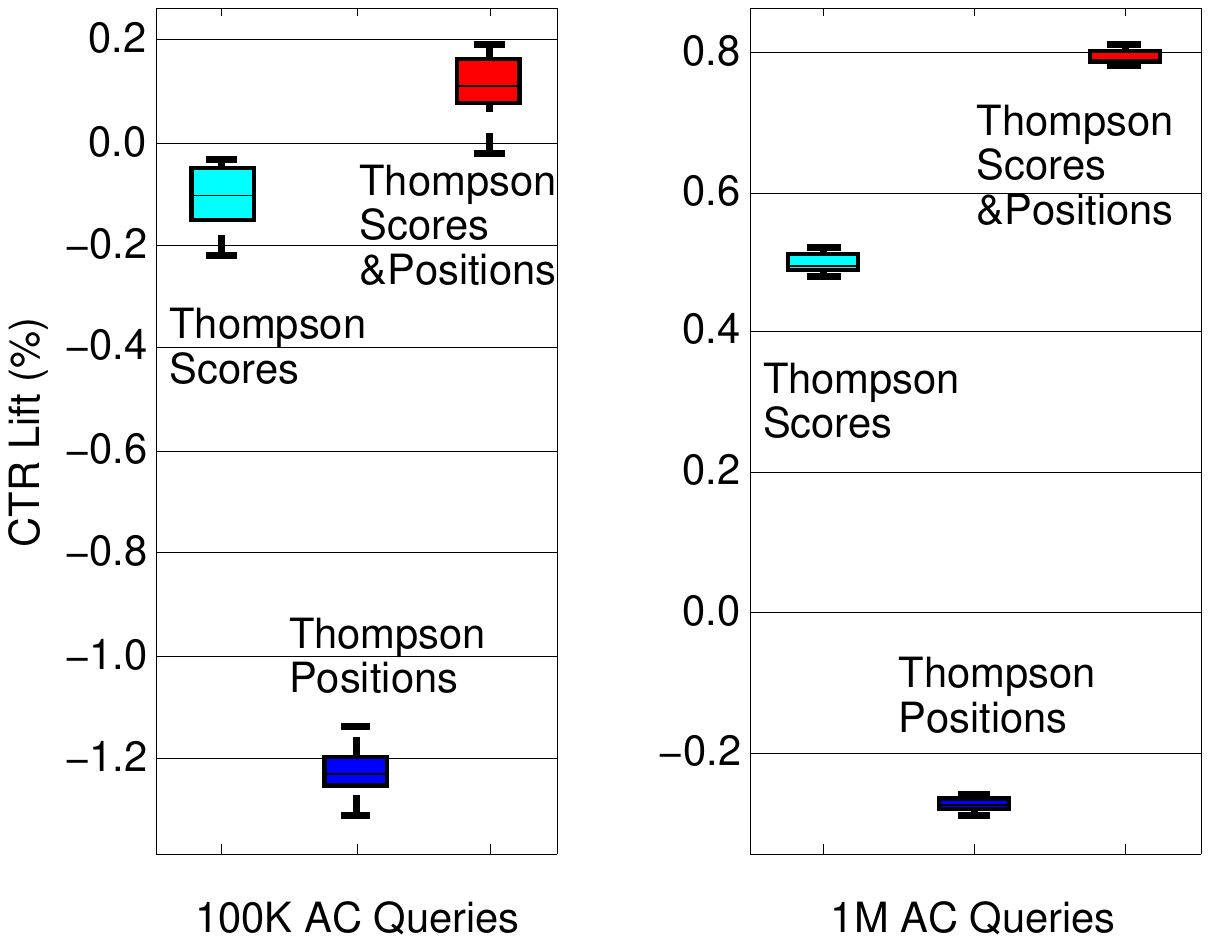}%
\caption{CTR lift during exploration for $k=2$ -  100K and 1M AC Queries.}
\label{fig:regretk2} 
\end{figure}

\textbf{Step 1}. We choose a setting for the results to be displayed by the offline system $k=2$ (Figures~\ref{fig:modelimprovementk2} and \ref{fig:regretk2}) and $k=3$ (Figures~\ref{fig:modelimprovementk3} and \ref{fig:regretk3}). 

\textbf{Step 2}. For the \emph{No Exploration} baseline we take the top $k$ suggestions and train the baseline model on them. If the production model is trained on only the top $k$ results from its logs it will achieve exactly the same performance as the \emph{No Exploration} model from the plots. 

\textbf{Step 3}. On the same set of queries, used to train the baseline model, we run the discussed Thompson sampling policies, with $n$ buckets as described in Section~\ref{sec:policies}, dividing the scores interval into 100 equal subintervals as mentioned in the same section. The policies involve a stochastic step - each of them decides randomly to place on position $i=k$ one of the suggestions from positions $i=k, k+1,...,N$, as described in Section~\ref{sec:simulation}. The resulting top $k$ suggestions are what a scaled down version of the production system would log if it utilizes the respective policy. We use this data to train the same type of model configured identically as the baseline model. Therefore, the improvement observed on the test period can be attributed entirely to the exploration data collected through the policies. 

\textbf{Step 4}.  If a query has fewer than, or equal to, $k$ results in the production logs we add it to the training set as is because in this case there are no results to explore from. We estimate that approximately 60\% of the queries have more than two results and approximately 50\% have more than three results. These are the queries which contribute to exploration for the two settings of $k=2$ and $k=3$ respectively.

\textbf{Step 5}. Finally, in training the models, we re-weight the examples collected with the exploration policies using the weighting schemes discussed in Section~\ref{sec:weighting}. 

\subsection{Evaluation of policies}\label{sec:evalpolicies}

\begin{figure}[!t]
\centering
	\includegraphics[width=3.2in,bb=133 250 480 530, clip]{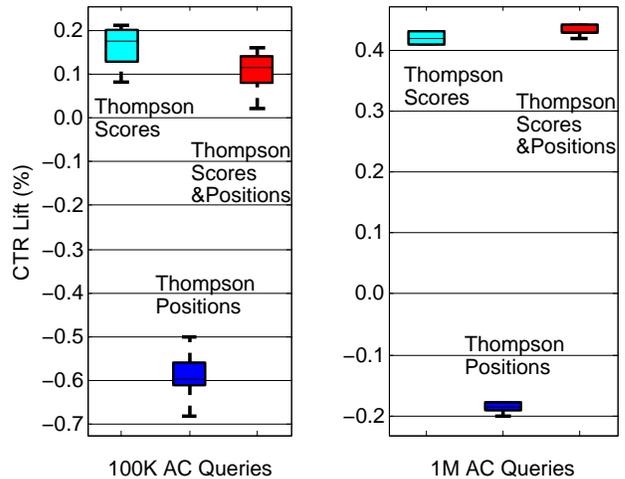}%
\caption{CTR lift during exploration for $k=3$ -  100K and 1M AC Queries.}
\label{fig:regretk3} 
\end{figure}

Let us now discuss in some more detail the results from the different experiments. 

\textbf{Model improvement}. 
Figures~\ref{fig:modelimprovementk2} and \ref{fig:modelimprovementk3} show the model improvement achieved with different policies for $k=2$ and $k=3$ suggestions respectively. CTR is measured as the fraction of queries that would result in a click over all queries from the test month in an Autocomplete system that would display up to $k$ suggestions. 

Naturally, having more data to train on improves the performance of all models. The CTR for all models for $k=2$ (Figure~\ref{fig:modelimprovementk2}) ranges from 0.87 to 0.92 while for $k=3$ it is between 0.94 and 0.96 (Figure~\ref{fig:modelimprovementk3}). The larger CTR for $k=3$ is due to two factors: 1) For $k=3$ the system displays more results which are more likely to contain the clicked suggestion from the production logs; 2) There is difference in the number of training examples too. For instance, for $k=2$ we still use 1M queries in the largest experiment, but each query contributes with up to two results, i.e. we have approximately 1.5M training examples in total. For $k=3$ there are approximately 2.5M examples for the same set of queries. 

We can see from the figures that in both cases the policies which account for bias in the model scores, \emph{Thompson over Scores} (\emph{Scores} for simplicity) and \emph{Thompson over Scores\&Positions} (\emph{Scores\&Positions} for simplicity), achieve large improvement over the production \emph{No Exploration} system. The advantage of the \emph{Scores} policy is especially striking for $k=2$ yielding a consistent 1.5\% model improvement across all query set sizes. In fact, using \emph{Scores} or \emph{Scores\&Positions} with just 50K training queries results in models which have better test performance for both settings of $k$ than the production \emph{No Exploration} model trained on 1M queries!

What is the reason for \emph{Scores} to perform better than \emph{Scores\&Positions}, and for both of them to outperform \emph{Positions} which barely improves on the production model for 1M queries? We believe the answer is twofold. 

First, there is a not too pronounced position bias associated with the problem. In web search often multiple results are relevant to a query. Showing them on different positions frequently yields different CTR. Here the intent is usually very well specified. Users are mostly interested in one particular geo-entity, e.g. a specific restaurant, park or a residential address. There are also only up to five suggestions per query which are easy to inspect visually. 

Second, there is very stable CTR associated with each position. Higher positions have higher CTR. Thompson sampling policies that utilize position information quickly identify this and start apportioning most of the exploration to the optimal position. Therefore, exploration stops too early.

\begin{figure}[!t]
\centering
	\includegraphics[width=3.3in,bb=103 250 495 560, clip]{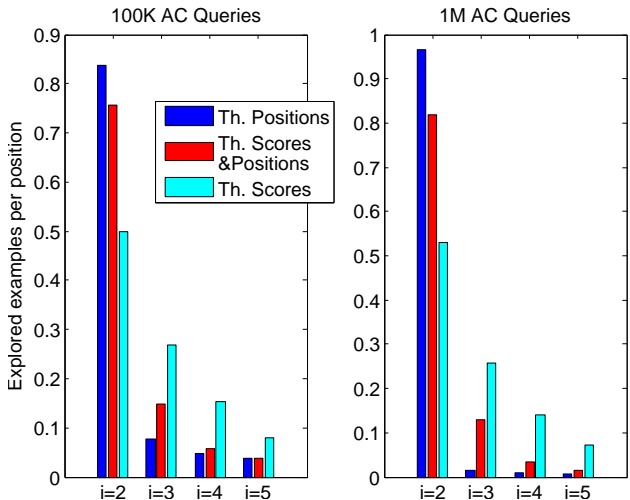}
\caption{Examples explored from different positions for $k=2$.}
\label{fig:exploredposk2} 
\end{figure}

\textbf{Exploration rate $\epsilon$}. 
For all experiments we use exploration rate $\epsilon=1$ (line 9, 11 in Algorithm~\ref{alg:thompson}). Decreasing $\epsilon$ forces Thompson sampling to continue exploration longer. As we mentioned above, \emph{Positions} stops exploration too early. To force further exploration, for this policy only, we decrease the rate to $\epsilon=0.01$. Still, as we see in Figures~\ref{fig:modelimprovementk2} and~\ref{fig:modelimprovementk3} \emph{Positions} only marginally improves on \emph{No Exploration} for 1M examples. It should be noted that one cannot decrease $\epsilon$ dramatically because then there is significant drop in CTR during exploration, which might be too steep of a price to pay for model improvement.

\textbf{Weighting scheme}.
We repeated all experiments from Figures~\ref{fig:modelimprovementk2} and~\ref{fig:modelimprovementk3} with both weighting schemes discussed in Section~\ref{sec:weighting}. Though propensity based weighting is considered a more suitable weighting scheme in the literature~\cite{Li_UnbiasedOfflineBandit2011}, for our multi-result ranking problem we observed a different outcome. For $k=2$ the multinomial weighting produced over 0.5\% improvement compared to propensity weights. Multinomial weighting was also slightly better, though less pronounced, in the case of $k=3$. All model improvement results presented here reflect multinomial weighting. 

\textbf{CTR lift in exploration}.
So far we saw that we can train models that outperform the production one by using exploration data from specific policies. We now look into what price the production system will pay for this model improvement. The common understanding is that as EE selects potentially suboptimal results production systems should always sustain a drop in CTR during the period of exploration. 

Figure~\ref{fig:regretk2} and~\ref{fig:regretk3}  demonstrate one of the major contributions of this work. They show that the above assumption is not necessarily true. On the contrary, if the right policy is selected multi-result ranking systems can even record a lift in their CTR. Here, \emph{CTR lift during exploration}  is defined as the CTR of each policy minus the CTR of the production \emph{No Explore} system both computed during the training month. 

The \emph{Positions} policy indeed impacts the CTR of the production system negatively: 1.2\% and 0.3\% drop in CTR for $k=2$, 100K and 1M queries Figure~\ref{fig:regretk2}, and 0.6\% and 0.2\% drop in CTR for $k=3$ Figure~\ref{fig:regretk3}. The results are for $\epsilon=0.01$, yet even for $\epsilon=1$ we observe a drop in CTR. 

\begin{figure}[!t]
\centering
	\includegraphics[width=3.3in,bb=103 250 495 560, clip]{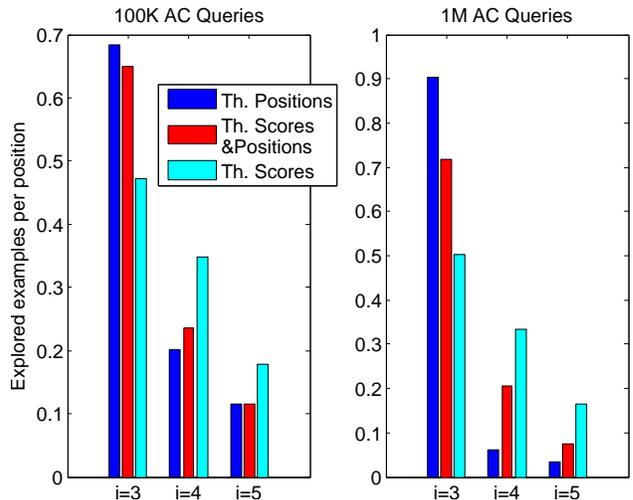}
\caption{Examples explored from different positions for $k=3$.}
\label{fig:exploredposk3} 
\end{figure}

With the \emph{Scores}  and \emph{Scores\&Positions} policies, however, we observe stable lift in CTR especially as the query set size grows to 1M. For $k=2$ and dataset size of 1M queries the policies improve the production CTR with 0.5\% and 0.8\% Figure~\ref{fig:regretk2}. For $k=3$ the lift is approximately 4.2\% and 4.3\% Figure~\ref{fig:regretk3}. The reason that for $k=3$ the lift is lower is due to the fact that in this setting there is a smaller pool of candidates for the system to explore --- it runs exploration among only three positions $k=3$, $k=4$ and $k=5$. 

Finally, Figure~\ref{fig:exploredposk2} and Figure~\ref{fig:exploredposk3} show from which positions the selection was performed during exploration in the case of $k=2$ and $k=3$ respectively. As can be seen, \emph{Positions} is very conservative and explores mostly the optimal position $i=2$ in Figure~\ref{fig:exploredposk2} and $i=3$ in Figure~\ref{fig:exploredposk3}, when the number of examples increases to 1M, even though we have set $\epsilon=0.01$. Another interesting observation is that \emph{Scores\&Positions} is more conservative, selecting fewer examples from suboptimal positions. As we saw above, this leads to a greater CTR lift during exploration than \emph{Scores}. However, it also produces less of an improvement in the ranking model.

%%%%%%%%%%%%%%%%%%%%%%%%%%%%%%%%%%%%%%%%%%%%%%%%%%%%%%%%%%%%%%%%%
% Section: Related work
%%%%%%%%%%%%%%%%%%%%%%%%%%%%%%%%%%%%%%%%%%%%%%%%%%%%%%%%%%%%%%%%%
\section{Related work}\label{sec:relatedwork}
Explore-exploit techniques hold the promise of improving machine-learned models by collecting high-quality, randomized data. A common concern in production teams, however, is that they may invest resources integrating EE in their systems, suffer high cost during the EE period, and end up with data that does not lead to substantially better models. To address this concern multiple efforts have focused on building offline systems that try to quantify a priori the effects that EE will have on the system~\cite{Bottou_Counterfactual2013,Langford_Scavenging2008,Li_ContextualBandidNews2010, Li_UnbiasedOfflineBandit2011, Strehl_LoggedData2011,Li14Counterfactual}. All of these works focus on how data collected with a production model or another policy~\cite{Langford_Scavenging2008} can be used to estimate a priori the performance of a new policy. Under the assumption of stationary data distribution, it can be proved that weighting data inversely proportional to the propensity scores leads to unbiased offline estimators, i.e. models for which we can provide guarantees will behave in a certain way in production~\cite{Bottou_Counterfactual2013,Strehl_LoggedData2011}. Many recent works adopt the evaluation approach (e.g., \cite{Bottou_Counterfactual2013,Li_UnbiasedOfflineBandit2011}). While theoretically sound, these offline frameworks make a few assumptions which may not always be present. For instance, as we noted earlier, in certain cases the propensity scores can not be computed in closed form, e.g. in Thompson sampling~\cite{Cook_ProbInequalities2008}, so one needs to use approximations as the one implemented here. Another problem lies in the fact that it is often impractical to assume stationary distribution --- for instance we find in our data many seasonal queries, queries which result from hot geo-political news etc, all of which impact the CTR of the system significantly. 

Our work can be considered a special case of the generic \emph{Contextual Bandid} framework  \cite{Barto85Pattern,Langford_EpochGreedy2008,Li_ContextualBandidNews2010,Wang05arbitraryside}. Unlike these works where the context is assumed to come in the form of additional observations~\cite{Wang05arbitraryside} or features, e.g. personalized information \cite{Li_ContextualBandidNews2010}, the context in our case is in the rich structure of the problem. For example, most of the works mentioned in this section focus on the single result case, i.e. there are $k$-arms to choose from but ultimately only one result is displayed to the user. We focus on multi-result ranking systems instead. As we saw here many real-world problems follow the multi-result ranking settings, which require special handling due to presence of position and ranking score bias. More recently~\cite{Kale11Nonstochastic} has discussed the multi-result setting. The authors describe a non-stochastic procedure for optimizing a loss function which is believed to lead to proper exploration. While the work is theoretically sound it does not show whether the approach leads to improvement of the underlying model. The method also lacks some of the observed convergence properties of Thompson sampling which starts apportioning examples to buckets which it overtime finds likely to lead to higher CTR.

The effectiveness of Thompson sampling has been noted previously by~\cite{Graepel10Web,Scott10Modern,Chapelle_Thompson11} and others. Subsequently efforts have focused on understanding better the theoretical properties of the algorithm (e.g., \cite{Kaufmann_ThompsonAnalysis2012}) leaving aside the important implementation considerations which we raised, namely, that in the context of multi-result ranking there are multiple ways to define the buckets (arms), and that different definitions lead to different semantic interpretation and different results.

\section{Conclusion}
We presented an offline framework which allows evaluation of EE policies prior to their deployment in an online environment. The framework allowed us to define and compare several different policies based on Thompson sampling. We demonstrated an interesting effect with significant practical implications. Contrary to the common belief, that a production system often has to pay a price of (possibly steep) CTR decrease during exploration, we show that the opposite can happen. If implemented suitably, a Thompson sampling policy can, in fact, have twofold benefits: it can collect data that improves the baseline model performance significantly and at the same time it can lift the CTR of the production system during the period of exploration. 

\bibliographystyle{abbrv}
\bibliography{mybibliography}{}

\begin{thebibliography}{10}

\bibitem{Barto85Pattern}
A.~G. Barto and P.~Anandan.
\newblock Pattern-recognizing stochastic learning automata.
\newblock {\em IEEE Transactions on Systems, Man, and Cybernetics},
  15(3):360--375, 1985.

\bibitem{Bottou_Counterfactual2013}
L.~Bottou, J.~Peters, J.~Qui\~{n}onero Candela, D.~X. Charles, D.~M.
  Chickering, E.~Portugaly, D.~Ray, P.~Simard, and E.~Snelson.
\newblock Counterfactual reasoning and learning systems: The example of
  computational advertising.
\newblock {\em J. Mach. Learn. Res.}, 14(1):3207--3260, Jan. 2013.

\bibitem{ChapelleLearningToRank}
O.~Chapelle and Y.~Chang.
\newblock Yahoo! learning to rank challenge overview.
\newblock {\em Journal of Machine Learning Research - Proceedings Track}, pages
  1--24, 2011.

\bibitem{Chapelle_Thompson11}
O.~Chapelle and L.~Li.
\newblock An empirical evaluation of thompson sampling.
\newblock In {\em Advances in Neural Information Processing Systems 24}, pages
  2249--2257, 2011.

\bibitem{Cook_ProbInequalities2008}
J.~D. Cook.
\newblock Numerical computation of stochastic inequality probabilities.
\newblock {\em UT MD Anderson Cancer Center Department of Biostatistics Working
  Paper Series}, Working Paper 46, 2008.

\bibitem{Friedman00greedyfunction}
J.~H. Friedman.
\newblock Greedy function approximation: A gradient boosting machine.
\newblock {\em Annals of Statistics}, 29:1189--1232, 2000.

\bibitem{Graepel10Web}
T.~Graepel, J.~Q. Candela, T.~Borchert, and R.~Herbrich.
\newblock Web-scale {Bayesian} click-through rate prediction for sponsored
  search advertising in {Microsoft's Bing} search engine.
\newblock In {\em Proceedings of the Twenty-Seventh International Conference on
  Machine Learning}, ICML'10.

\bibitem{Kale11Nonstochastic}
S.~Kale, L.~Reyzin, and R.~E. Schapire.
\newblock Non-stochastic bandit slate problems.
\newblock In {\em Advances in Neural Information Processing Systems NIPS-2010},
  pages 1054--1062, 2011.

\bibitem{Kaufmann_ThompsonAnalysis2012}
E.~Kaufmann, N.~Korda, and R.~Munos.
\newblock Thompson sampling: An asymptotically optimal finite-time analysis.
\newblock In {\em Proceedings of the 23rd International Conference on
  Algorithmic Learning Theory}, ALT'12, pages 199--213. Springer-Verlag, 2012.

\bibitem{Langford_Scavenging2008}
J.~Langford, A.~Strehl, and J.~Wortman.
\newblock Exploration scavenging.
\newblock In {\em Proceedings of the 25th International Conference on Machine
  Learning}, ICML '08, pages 528--535, New York, NY, USA, 2008.

\bibitem{Langford_EpochGreedy2008}
J.~Langford and T.~Zhang.
\newblock The epoch-greedy algorithm for multi-armed bandits with side
  information.
\newblock In J.~Platt, D.~Koller, Y.~Singer, and S.~Roweis, editors, {\em
  Advances in Neural Information Processing Systems 20}, pages 817--824. 2008.

\bibitem{Li14Counterfactual}
L.~Li, S.~Chen, A.~Gupta, and J.~Kleban.
\newblock Counterfactual analysis of click metrics for search engine
  optimization.
\newblock Technical report, Microsoft Research, 2014.

\bibitem{Li_ContextualBandidNews2010}
L.~Li, W.~Chu, J.~Langford, and R.~E. Schapire.
\newblock A contextual-bandit approach to personalized news article
  recommendation.
\newblock In {\em Proceedings of the 19th International Conference on World
  Wide Web}, WWW '10, pages 661--670, New York, NY, USA, 2010.

\bibitem{Li_UnbiasedOfflineBandit2011}
L.~Li, W.~Chu, J.~Langford, and X.~Wang.
\newblock Unbiased offline evaluation of contextual-bandit-based news article
  recommendation algorithms.
\newblock In {\em Proceedings of the 4th ACM International Conference on Web
  Search and Data Mining}, WSDM '11, pages 297--306, 2011.

\bibitem{NiculescuMizilCalibration2005}
A.~Niculescu-Mizil and R.~Caruana.
\newblock Predicting good probabilities with supervised learning.
\newblock In {\em Proceedings of the 22Nd International Conference on Machine
  Learning}, ICML '05, pages 625--632, 2005.

\bibitem{Scott10Modern}
S.~L. Scott.
\newblock A modern {Bayesian} look at the multi-armed bandit.
\newblock {\em Applied Stochastic Models in Business and Industry},
  26:639--658, 2010.

\bibitem{Strehl_LoggedData2011}
A.~Strehl, L.~Li, J.~Langford, and S.~M. Kakade.
\newblock Learning from logged implicit exploration data.
\newblock In {\em NIPS}, pages 2217--2225, 2011.

\bibitem{Thompson33Likelihood}
W.~R. Thompson.
\newblock On the likelihood that one unknown probability exceeds another in
  view of the evidence of two samples.
\newblock {\em Biometrika}, 25(3--4):285--294, 1933.

\bibitem{Wang05arbitraryside}
C.~Wang, S.~R. Kulkarni, and H.~V. Poor.
\newblock Arbitrary side observations in bandit problems.
\newblock {\em Adv. Applied Math}, 34:903--936, 2005.

\bibitem{WuBoostingIR2010}
Q.~Wu, C.~J. Burges, K.~M. Svore, and J.~Gao.
\newblock Adapting boosting for information retrieval measures.
\newblock {\em Inf. Retr.}, 13(3):254--270, June 2010.

\end{thebibliography}
% ACM needs 'a single self-contained file'!
%\balancecolumns
\balancecolumns
% That's all folks!
\end{document}